\RequirePackage{fix-cm}
\RequirePackage{amsmath}     

\documentclass[a4paper,10pt]{article}

\usepackage{graphicx}
\usepackage{url}
\usepackage{algorithm}
\usepackage{algorithmic}
\usepackage{authblk}

\title{Distributed ReliefF based Feature Selection in Spark\footnote{This is a pre-print of an article published in Knowledge And Information Systems. The final authenticated version is available online at: https://doi.org/10.1007/s10115-017-1145-y}}

\author[1]{Raul-Jose Palma-Mendoza}
\author[2]{Daniel Rodriguez}
\author[2]{Luis de-Marcos}

\urldef{\mailsNAUH}\path|raul.palma@unah.edu.hn|
\urldef{\mailsUAH}\path|{daniel.rodriguezg, luis.deMarcos}@uah.es|

\affil[1]{Systems Engineering Department\\
National Autonomous University of Honduras, Tegucigalpa, Honduras
\mailsNAUH
}

\affil[2]{University of Alcala, 28871 Alcala de Henares, Madrid, Spain
\mailsUAH
}

\date{}

\begin{document}

\maketitle

\begin{abstract}
Feature selection (FS) is a key research area in the machine learning and data mining fields, removing irrelevant and redundant features usually helps to reduce the effort required to process a dataset while maintaining or even improving the processing algorithm's accuracy. However, traditional algorithms designed for executing on a single machine lack scalability to deal with the increasing amount of data that has become available in the current Big Data era. ReliefF is one of the most important algorithms successfully implemented in many FS applications. In this paper, we present a completely redesigned distributed version of the popular ReliefF algorithm based on the novel Spark cluster computing model that we have called DiReliefF. Spark is increasing its popularity due to its much faster processing times compared with Hadoop's MapReduce model implementation. The effectiveness of our proposal is tested on four publicly available datasets, all of them with a large number of instances and two of them with also a large number of features. Subsets of these datasets were also used to compare the results to a non-distributed implementation of the algorithm. The results show that the non-distributed implementation is unable to handle such large volumes of data without specialized hardware, while our design can process them in a scalable way with much better processing times and memory usage.\\


\end{abstract}

\section{Introduction}
\label{intro}

In the last years we have witnessed a vast increase in the amount of data that is being stored and processed by organizations of all types. As stated by Xindong et al.~\cite{XindongWu2014}, the so-called big data revolution has come to us not only with many challenges but also with plenty of opportunities that these organizations are willing to embrace. According to Rajamaran and Ulman~\cite{rajaraman2014mining}, the main challenge is to extract useful information or knowledge from these huge volumes of data that enable us to predict or to better understand the phenomena involved in its generation. These tasks are commonly tackled using data mining. 

As part of the data mining process, feature selection or feature subset selection is a crucial preprocessing step for identifying the most relevant attributes from a dataset. Removing irrelevant and redundant attributes not only can generate less complex and more accurate models but also a reduced dataset allows us to enhance the performance of many data mining schemes. Feature selection algorithms are usually classified into three types: \emph{wrappers}, \emph{filters} and \emph{embedded methods}. Wrappers refer to methods that require the learning of a classifier based on a single or on a subset of the original features, they are usually more computationally expensive. Filter methods rely only on the characteristics of data and are independent of any learning scheme, thereby requiring less computational effort. Finally, embedded methods refer to those techniques where the selection of features is carried out as part of the classification process. 

Another important feature selection classification, as stated by Garc\'ia et al.~\cite{García2015}, is the one that comes from viewing the feature selection as a search problem with the aim of finding a high quality feature subset, and thereby, they classify the methods according to the type of search performed: \emph{exhaustive}, \emph{heuristic} and \emph{stochastic}. However, as not all feature selection techniques can be viewed as these types of search, a fourth category needed to be included, \emph{feature weighting}.

In this context, ReliefF~\cite{Kononenko1994} is a widely applied feature weighting technique that estimates the quality of the features from a given dataset by assigning weights to each of them. It can be used as filter feature selection method by defining a significance threshold and selecting features with quality above it. As a result of its advantages such as being able to work with nominal or continuous features, handling multi-class problems, detecting features interactions, handling missing data and noisy tolerance, it is considered one of the most convenient filter-based feature selection methods available \cite{Bolón-Canedo2012}.

On the other hand, after its 2004 seminal paper, Google's MapReduce~\cite{Dean2004a,Dean2008} emerged as a programing model that simplified the development of scalable applications that process and generate large scale datasets. These applications are defined in terms of map and reduce tasks that are automatically parallelized by a programming framework and executed in clusters of tens or even thousands of machines, handling failures and scheduling resources. One important fact is that MapReduce was designed to run on commodity hardware, because it leads to lower costs per processor and per unit of memory of the cluster. The standard MapReduce implementation to date is Hadoop MapReduce~\cite{hadoop}, an open source implementation mainly developed at Yahoo Labs. However, the framework uses disk writing between every MapReduce job with the objective of recovering from failures, and this becomes a bottleneck for iterative nature algorithms like the ones used in machine learning and data mining, including the original ReliefF algorithm. For that reason, Spark~\cite{Zaharia2010} has gained much attention in the last couple of years, since it presents an improved model that is capable of handling most of its operations in-memory while maintaining the fault tolerance and scalability of MapReduce.

In this paper we present DiReliefF\footnote{\url{https://github.com/rauljosepalma/DiReliefF}}, a distributed and scalable redesign of the original ReliefF algorithm based on the Spark computing model, enabling it to deal with much larger datasets in terms of both instances and features than the traditional version would be able to handle. In addition, after presenting and discussing the design of the algorithm we compare several runs of it with the traditional version implemented on the WEKA \cite{Hall2009} platform, using 4 open datasets with numbers of instances in the order of $10^7$ and a number of features that ranges in the order of $10^1$ until $10^3$. The main conclusion obtained is that is practically unfeasible to deal with such high amounts of data using the traditional version on standard hardware, while the distributed version is able to handle them smoothly.

\section{Related Work}
\label{sec:relatedWork}
ReliefF was first described by Kononenko in 1994 \cite{Kononenko1994} and since then, many applications, extensions and improvements have been published. Moreover, ReliefF is itself an extension of the original Relief algorithm developed by Kira and Rendell \cite{Kira1992}, the latter was initially limited to binary class problems while the former can handle multi-class problems. As recent examples of those extensions, Reyes et al.~\cite{Reyes2015} presented three of them for multi-label problems and compared them with previous extensions for the same purpose. Zafra et al. \cite{Zafra2012} extended ReliefF to the problem of multi-instance learning. Greene et al. \cite{Greene2009} proposed an adaptation to enhance ReliefF's ability to detect feature interactions called SURF (Spatially Uniform ReliefF) and applied it in the genetic analysis of human diseases.

One of ReliefF's mayor flaws is its incapacity to detect redundant features, and hereby some attempts have been made to overcome this flaw. For example, Li and He~\cite{Li2016} used a forward selection algorithm to select non redundant critical quality characteristics of complex industrial products. Zhang et al.~\cite{Zhang2008} combined ReliefF with the mRMR (minimal-redundancy-maximal-relevancy) algorithm \cite{Peng2005} to select non redundant gene subsets that best discriminate biological samples of different types. 

As it might be expected, most feature selection algorithms have asymptotic complexities that depend on the number of features or instances, and in our case, ReliefF depends linearly on both of them~\cite{Robnik2003}. Thereby, its performance gets compromised when faced with datasets with high dimensionality and/or high number of instances. For this reason, many attempts have been made to make feature selection methods, including ReliefF more scalable.

Recently, Canedo et al.~\cite{Bolon-Canedo2015} proposed a framework to deal with high dimensional data, distributing the dataset by features, processing in parallel the pieces, and then performing a merging procedure to obtain a single selection of features, however this method is oriented for high dimensional datasets and no tests were made with datasets with high amounts of instances. A some way similar approach was followed by Peralta et al.~\cite{Peralta2015}, who used the MapReduce model to implement a wrapper-based evolutionary search feature selection method. In this case, the data is split by instances, and an evolutionary feature selection is performed over each of these pieces. Furthermore, the reduce step basically uses a simple majority voting of the selected features with a user-defined threshold to select the definitive subset of features. All the tests were carried out with the EPSILON dataset, that we also use here (see Section~\ref{sec:experiments}).

Zhao et al.~\cite{Zhao2012} presented a distributed parallel feature selection method based on variance preservation using the SAS High-Performance Analytics \footnote{\url{http://www.sas.com/en_us/software/high-performance-analytics.html}} proprietary software. Experimental work was carried out with both high dimensional and high number of instances datasets.

Kubica et al.~\cite{Kubica2011} developed parallel versions of three forward search based feature selection algorithms, a wrapper with a logistic regression classifier is used to guide the search that is parallelized using the MapReduce model.

It is also worth mentioning the recent work of Wang et al.~\cite{Wang2016} that uses the Spark computing model to implement feature selection strategy for classification of network traffic. Their approach basically consists in two phases, first it generates a feature ranking based on the Fisher score, and then a forward search is done using a wrapper approach. Nevertheless, this method can only be applied to continuous data.

As it can be seen, none of the above contributions implements a purely filter approach, even in the case of the framework proposed by Bol\'on-Canedo et al.~\cite{Bolon-Canedo2015} that allows us to apply filters to parts of the dataset, the results are then merged using a wrapper. However, in a recent publication, Ram\'irez-Gallego et al.~\cite{Ramirez-Gallego2016} presented three implementations of an extended version of the popular mRMR feature selection filter, including a distributed version under Spark. 

Moreover, for the specific case of ReliefF, Huang et al.~\cite{Huang2009} proposed an optimization to improve the computation efficiency on large datasets, but this improvement is only useful when no random sampling of the instances is performed for the weights approximation. In other words, it only works when the $m$ parameter (see Section~\ref{sec:reliefF}) is set equal to the number of instances. Furthermore, the experimental work they performed was carried out with datasets with a number of instances in the order of $10^4$, which is still manageable by a single machine.

Finally, to the best of our knowledge, no published contribution has ever attempted to redesign the ReliefF filter method to a distributed environment as is proposed here.

\section{ReliefF}
\label{sec:reliefF}

We briefly describe the original ReliefF algorithm by Kononenko \cite{Kononenko1994} that will serve as a conceptual basis for the redesign presented in Section~\ref{sec:direliefF}. ReliefF's central idea consists in evaluating the quality of the features by their ability to distinguish instances from one class to another in a local neighborhood, i.e. the best features are those that contribute more to increase distance between different class instances while contribute less to increase distance between same class instances. ReliefF, as mentioned above, is an extension of the original Relief method, but it is capable of working with multi-class, noisy and incomplete datasets.

\begin{algorithm}
\caption{ReliefF~\cite{Robnik2003,Kononenko1994}}
\label{alg:reliefF}
\begin{algorithmic}[1] 
\STATE calculate prior probabilities $P(C)$ for all classes
\STATE set all weights $W[A] := 0.0$
\FOR{$i = 1$ \TO $m$}
  \STATE randomly select an instance $R_i$
  \STATE find $k$ nearest hits $H_j$
  \FORALL{classes $C \neq cl(R_i)$}
    \STATE from class $C$ find $k$ nearest misses $M_j(C)$
  \ENDFOR
  \FOR{$A := 1$ \TO $a$}
    \STATE $AH := - \sum_{j = 1}^{k} \mathit{diff}(A,R_i,H_j) / (m \cdot k)$ \label{alg:diff1}
    \STATE $AM := \sum_{C \neq cl(R_i)} \left [ \left ( \frac{P(C)}{1 - P(cl(R_i))} \right )\sum_{j = 1}^{k} \mathit{diff}(A,R_i,M_j(C))  \right ] / (m \cdot k)$ \label{alg:diff2} 
    \STATE $W[A] := W[A] + (AH + AM)$
  \ENDFOR
\ENDFOR
\end{algorithmic}
\end{algorithm}

Algorithm \ref{alg:reliefF} displays ReliefF's pseudo-code, mostly preserving the original notation used in \cite{Robnik2003}. As we can observe, it consists of a main loop that iterates $m$ times, where $m$ corresponds to the number of samples from data to perform the quality estimation. 
Each selected sample $R_i$ equally contributes to the $a$-size weights vector $W$, where $a$ is the number of features in the dataset. The contribution for the $A$-esim feature is calculated by first finding $k$ nearest neighbors of the actual instance for each class in the dataset. The $k$ neighbors that belong to the same class as the actual instance are called hits($H$), and the other $k \cdot (c - 1)$ neighbors are called misses ($M$), where $c$ is the total number of classes, and $cl(R_i)$, represents the class of the $i-$esim sample. Once neighbors are found, their respective contributions to $A$-esim feature are calculated. The contribution of the hits collection $AH$ is equal to the negative of the average of the differences between the actual instance and each hit, note that this is a negative contribution because only non desirable features should contribute to create differences between neighbor instances of the same class. Analogously, the contribution of the misses collection $AM$ is equal to the weighted average of the differences between the actual instance and each miss, this is of course a positive contribution because good features should help to differentiate between instances of a different class. The weights for this summation are defined according to the prior probability of each class, calculated from the dataset. Finally, it is worth mentioning that dividing both $AH$ and $AS$ by $m$ simply indicates another average between the contributions of all $m$ samples. Since the $\mathit{diff}$ function returns values between $0$ and $1$, the ReliefF's weights will lie in the range $[-1, 1]$, and must be interpreted in the positive direction: the higher the weight, the higher the corresponding feature's relevance.

The $\mathit{diff}$ function is used in two cases in the ReliefF algorithm, the obvious one is between lines \ref{alg:diff1} and \ref{alg:diff2} to calculate the weight. It is also used to find distances between instances, defined as the sum of the differences over every feature (Manhattan distance). The original $\mathit{diff}$ function used to calculate the difference between two instances $I_1$ and $I_2$ for a specific feature $A$ is defined in~(\ref{eq:diffNom}) for nominal features, and as in~(\ref{eq:diffNumOrig}) for numeric features. However, the latter has been proved to cause an underestimation of numeric features with respect to nominal ones in datasets with both types of features. Thereby, a so-called ramp function, depicted in~(\ref{eq:diffNumRamp}), has been proposed by Hong~\cite{Hong1997} to deal with these problem. The idea behind is to relax the equality comparison on~(\ref{eq:diffNumOrig}) by using two thresholds: $t_{eq}$ is the maximum distance between two features to still consider them equal, and analogously, $t_{diff}$ is the minimum distance between two features to still consider them different. Their default values are set to $5\%$ and $10\%$ of the feature's value interval respectively. In addition, there are also versions of the $\mathit{diff}$ function to deal with incomplete data, however, since the datasets chosen for the experiments in this work do not have missing values, we do not consider them here.

\begin{equation}
\label{eq:diffNom}
\mathit{diff}(A,I_1,I_2) = \begin{cases}
0 & \text{if } value(A, I_1) = value(A, I_2),\\
1 & \text{otherwise }
\end{cases}
\end{equation}

\begin{equation}
\label{eq:diffNumOrig}
\mathit{diff}(A,I_1,I_2) = \frac{\left | value(A, I_1) - value(A, I_2) \right |}{ max(A) - min(A)}
\end{equation}

\begin{equation}
\label{eq:diffNumRamp}
\mathit{diff}(A,I_1,I_2) = \begin{cases}
0 & \text{if } d \leq t_{eq}, \\
1 & \text{if } d > t_{\mathit{diff}}, \\
\frac{d-t_{eq}}{t_{\mathit{diff}}-t_{eq}} & \text{if } t_eq < d \leq t_{\mathit{diff}}
\end{cases}
\end{equation}

\section{Spark Cluster Computing Model}

We now briefly describe the main concepts behind the Spark computing model, focusing on those that will complete the conceptual basis for the description of our proposal in Section~\ref{sec:direliefF}. We also provide a short comparison with other existent computing models such as MapReduce, with the aim of justifying our selection of Spark. 

The main concept behind the Spark model is the Resilient Distributed Dataset or in short RDD. Zaharia et al.~\cite{Zaharia2010,Zaharia2012} defined an RDD as a read-only collection of objects, that is partitioned and distributed across the nodes of a cluster. It has the ability to automatically recover lost partitions through the record of lineage that knows the origin of the data and optionally, the calculations that went through it. Even more relevant is the fact that the operations run for an RDD are automatically parallelized by the Spark engine, this abstraction frees the programmer of having to deal with threads, locks and all the complexities involved in the traditional parallel programming.

There are two types of operations that can be executed on an RDD: (i) actions and (ii) transformations. On the one hand, \emph{actions} are the mechanism that permit to obtain results from a Spark cluster; five commonly used actions are: $reduce$, $sum$, $aggregate$, $sample$ and $collect$. The action $reduce$ is used to aggregate the elements of an RDD, by applying a commutative and associative function that receives as arguments two elements of the RDD an returns one element of the same type. Action $sum$ is simply a shorthand for a reduce action that sums all the elements on the RDD. Next, action $aggregate$ has a similar behavior to $reduce$, but its return type can be different from the type of the elements of the RDD. It works in two steps: the first one aggregates the elements of each partition and returns an aggregated value for each of them, the second one, merges these values between all partitions to a single one, that becomes the definitive result of the action. Lastly, actions $sample$ and $collect$ are also similar, the former takes an amount of elements and returns a random sample of this size from the RDD, and the latter, simply returns an array with all the elements in the RDD, this of course has to be done with care to prevent exceeding the maximum memory available at the driver (as later explained).

On the other hand, \emph{transformations} are the mechanism for creating an RDD from another one. Since RDDs are read-only, a transformation does not affect the original RDD but creates a new one. Three of the most important transformations are: $map$, $flatMap$ and $filter$. The first two: $map$ and $flatMap$ are similar, they return a new RDD that is the result of applying a function to each element of the original one. In the case of $map$, the function applied takes a single argument and returns a single element, thus the new RDD has the same number of elements that the original one. In the case of $flatMap$, the applied function takes a single element but it can return zero or more elements, therefore the resulting RDD is not required to have the same number of elements as the original one. Finally, $filter$ is straightforward, it receives a boolean function to discriminate dataset elements to return a subset of it.

\begin{figure*}
  \includegraphics[width=1\textwidth]{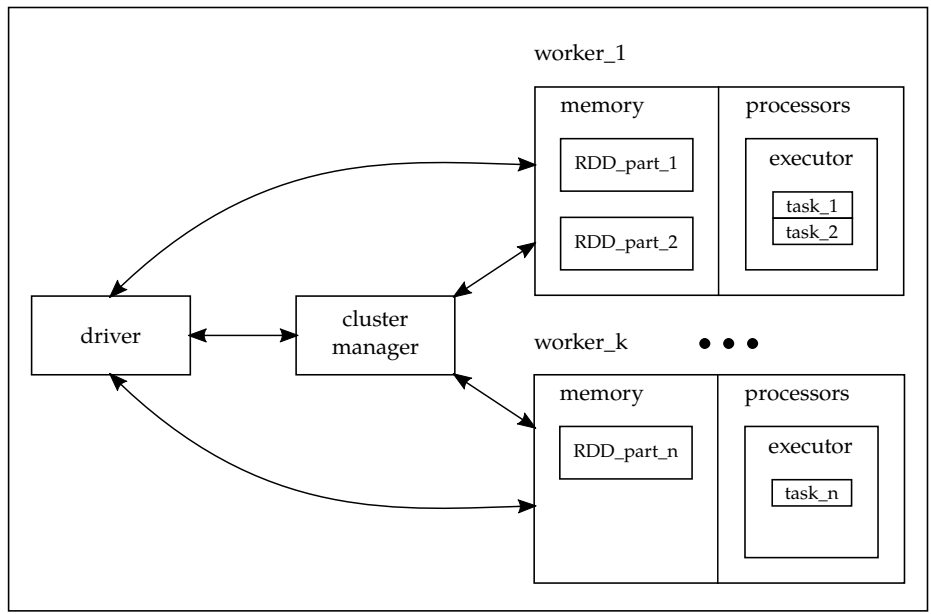}
\caption{Spark Cluster Architecture}
\label{fig:sparkArch}
\end{figure*}

In respect to the cluster's architecture, Spark follows the master-slave model. There is a cluster manager (master) through which the so-called driver program can access the cluster. The driver coordinates the execution of a user application by assigning tasks to the executors, which are programs that run in worker nodes (slaves). By default only one executor is run per worker. With regard to the data, RDD partitions are distributed across the worker nodes, and the number of tasks launched by the driver for each executor will be according to the number of partitions of the RDD residing in the worker. A detailed view of the discussed architecture with respect to the physical nodes can be seen in Figure~\ref{fig:sparkArch}.

\subsection{Other Cluster Computing Models}
Spark has quickly emerged from other cluster computing models (e.g. MapReduce as the most relevant) as the prefered platform due to its advantages. First of all, Spark was designed from the beginning to efficiently handle iterative jobs in memory, such as the ones used by many data mining schemes. This has lead to the quick development of a machine learning library~\cite{Meng2015} that contains redesigned distributed algorithms such as the one in this work. Moreover, besides the Spark author's own comparison \cite{Zaharia2010,Zaharia2012}, others \cite{Shi2015} have shown that Spark is faster than MapReduce in most of the data analysis algorithms tested. Second, any MapReduce program can be directly translated to Spark, i.e. the MapReduce model can be completely expressed using the $flatMap$ and $groupByKey$ operations in Spark. Finally, as previously stated, Spark provides a wider range of operations other than $flatMap$ and $groupByKey$.

This being said, its worth mentioning that other models have already tried to fulfill the lack of efficient iterative job handling in MapReduce. Two of them are HaLoop \cite{Bu2010} and Twister \cite{Ekanayake2010}.  However, even though they have support for executing iterative MapReduce jobs, automatic data partitioning, and Twister has the ability to keep it in-memory, they both prevented an interactive data mining and can indeed be considered subsets of Spark functionality. By the time, both projects have become outdated with newest versions dating from 2012 and 2011 respectively.

A recently work by Liu et al~\cite{Liu2016} compared parallelized versions of a neural network algorithm over Hadoop, HaLoop and Spark, concluding that Spark was the most efficient in all cases.

\section{DiReliefF}
\label{sec:direliefF}
In this section, we describe our proposed algorithm. The first design decision is where to concentrate the parallelization effort as the ReliefF algorithm could be described as an embarrassingly parallel algorithm since its outermost loop goes through completely independent iterations, each of these can be directly executed in different threads as stated by Robnik and Kononenko~\cite{Robnik2003}. However, parallelizing the algorithm in such way ties the parallelization to the number of samples $m$, and prevents Spark from doing optimizations based on the resources available, the size of the dataset, the number of partitions and the data locality. This would also require that every thread would read through the whole dataset, while as we show below, there is only one pass needed to process the distances and calculate the feature weights. Furthermore, continuing with Robnik and Kononenko discussion, ReliefF algorithm's complexity is $\mathcal{O}(m \cdot n \cdot a)$, where $n$ is the number of instances in the dataset, $m$ is the number of samples taken from the $n$ instances and $a$ is the number of features. Moreover, the most complex operation is the selection of the $k$ nearest neighbors for two reasons: first, the distance from the current sample to each of the instances must be calculated with $\mathcal{O}(n \cdot a)$ steps; and second, the selection must be carried out in $\mathcal{O}(k \cdot log(n))$ steps. As a result, the parallelization is focused on these stages rather than on the $m$ independent iterations.

We need to bear in mind that the ReliefF algorithm can be considered as a function applied to a dataset $DS$, having as input parameters the number of samples $m$ and the number of neighbors $k$, and returning an $a-$size vector of weights $W$, as shown in~(\ref{eq:reliefFFunction}). Thus, the ReliefF algorithm can be interpreted as the calculation of each individual weight $W[A]$, using~(\ref{eq:weightA}), where $\mathit{sdiffs}$~(\ref{eq:sumOfDiffs}) represents a function that returns the total sum of the differences in the $A-$esim feature between a given instance $R_i$, and a set $NN_{C,i}$ of $k$ neighbors of this instance where all belong to the same class $C$.

\begin{equation}
\label{eq:reliefFFunction}
\mathit{reliefF}(DS,s,w) = W
\end{equation}

\begin{equation}
\label{eq:weightA}
W[A] = \frac{1}{m} \cdot \sum_{i = 1}^{m} \left [ -\mathit{sdiffs}(A, R_i, cl(R_i)) + \sum_{C \neq cl(R_i)} \left [ \left ( \frac{P(C)}{1 - P(cl(R_i))} \right )
\mathit{sdiffs}(A,R_i,C) \right ] \right ]
\end{equation}

\begin{equation}
\label{eq:sumOfDiffs}
\mathit{sdiffs}(A,R_i,C) = \frac{1}{k} \cdot \sum_{j = 1}^{k} \mathit{diff}(A,R_i,NN_{C,i,j})
\end{equation}

The dataset $DS$ can be defined (see~(\ref{eq:dataset})) as a set of $n$ instances each represented as a pair $I_i = (F_i,C_i)$ of a features vector $F_i$ and a class $C_i$. 

\begin{equation}
\label{eq:dataset}
DS = \left \{ (F_1,C_1),(F_2,C_2),\cdots,(F_n,C_n) \right \}
\end{equation}

Given the initial definitions and assuming that the features types (nominal or numeric) are stored as metadata, we first calculate the maximum and minimum values for all continuous features in the dataset. These values are needed by the $\mathit{diff}$ function (see~(\ref{eq:diffNumRamp}) and~(\ref{eq:diffNumOrig})). Our implementation, as the original version, uses~(\ref{eq:diffNom}) for nominal attributes and selects between~(\ref{eq:diffNumRamp}) and~(\ref{eq:diffNumOrig}) for numeric attributes via an initialization parameter. The task of finding maximum and minimum values is efficiently achieved applying a $reduce$ action with a function $fmax$ ($fmin$) that given two instances returns a third one containing the maximum (minimum) values for each continuous feature. This is shown for maximum values on~(\ref{eq:max}).

\begin{align}
\label{eq:max}
MAX & = DS.reduce(fmax) \nonumber \\
MAX & = (max(F_1[1],\cdots,F_n[1]), \cdots , max(F_1[a],\cdots,F_n[a]))
\end{align}

The next step consists in calculating the prior probabilities of all classes in $DS$, this is also a requirement of the $\mathit{diff}$ function. This values can be essentially obtained by the means of a $map$ and a $reduceByKey$ transformation, the former simply returns a dataset with all instance classes paired with a value of one, and the latter sums these ones using the class as a key, thereby obtaining a set of pairs $(C_i, count(C_i))$ containing the classes and the number of instances in $DS$ belonging to that class, which can simply be divided by $n$ to obtain the priors. Equations~(\ref{eq:priors}) depict the previous discussion, note that with the use of $collect$, $P$ is turned into a local array rather than an RDD.

\begin{align}
\label{eq:priors}
f(I) & = (cl(I),1) \nonumber \\
g(a,b) & = a + b \nonumber \\
h((a,b)) & = (a, b / n) \nonumber \\
P & = DS.map(f).reduceByKey(g).map(h).collect() \nonumber \\
P & = \left \{ (C_1, prior(C_1)), \cdots , (C_c, prior(C_c)) \right \}
\end{align}

A rather short step, is the selection of the $m$ samples, this is accomplished with the use of the $takeSample$ action, as shown in~(\ref{eq:samples}).

\begin{equation}
\label{eq:samples}
R = ( R_1,\cdots,R_m) = DS.takeSample(m)
\end{equation}

Now we can proceed to the computationally intensive step of finding the $k$ nearest neighbors of each sample for each class. We first find the distances from every sample $m$ to each of the $n$ instances. This can be directly accomplished by means of a $map$ transformation applied to $DS$ where for every instance $I$, a vector of distances from it to the $m$ samples is returned (as shown in~(\ref{eq:distancesDataset})).

\begin{align}
\label{eq:distancesDataset}
distances(I) & = (distance(I,R_1), \cdots ,distance(I,R_m)) \nonumber \\
f(I) & = (I, distances(I)) \nonumber \\
DD & = DS.map(f) \nonumber \\
DD & = \left \{ (I_1, distances(I_1)),\cdots (I_n, distances(I_n))) \right \} 
\end{align}

Next, as shown in~(\ref{eq:nearestNeighbors}), we are able to obtain the nearest neighbors $NN$ matrix by using an $aggregate$ action, where each element $NN_{C,i}$ of this matrix is a vector with the $k$ nearest neighbors of a sample $R_i$ belonging to a class $C$. 

As stated before, the $aggregate$ action has two steps. The first step is defined in the function $localNN$ that returns a local neighbors matrix $LNN$ for each partition of the RDD. This matrix, has a similar structure as the $NN$ matrix but each vector element is treated instead as a $k$-sized binary heap that is used to incrementally store the nearest neighbors found during the traverse of the local partition.

The second step of the $aggregate$ action is the merging of the local matrices, the defined function $mergeNN$ combines two local matrices by merging its individual binary heaps, keeping only the elements with shorter distances. Once both functions have been defined, a call to the $aggregate$ action can be performed, providing also an empty matrix $LNN$ structure (with empty heaps) so the $localNN$ can start aggregating neighbors to it. Lastly note that since the $NN$ matrix is obtained via an action, it is a local object and not an RDD. Calculating it concludes the complex step in the algorithm, step that has been fully implemented using parallel operations.

\begin{align}
\label{eq:nearestNeighbors}
NN & = \begin{bmatrix}
NN_{1,1} & \cdots  & NN_{1,m}\\ 
\vdots & \ddots  & \vdots \\ 
NN_{c,1} & \cdots & NN_{c,m} 
\end{bmatrix} \nonumber \\
NN_{C,i} & = (N_1,\cdots,N_k \mid \forall N \ cl(N) = C) \nonumber \\
localNN(I, LNN_{in}) & = LNN_{out}  \nonumber \\
mergeNN(LNN_a, LNN_b) & = LNN_{merged} \nonumber \\
NN & = DS.aggregate(emptyNN, localNN, mergeNN)
\end{align}

\begin{figure}
  \includegraphics[width=1\textwidth]{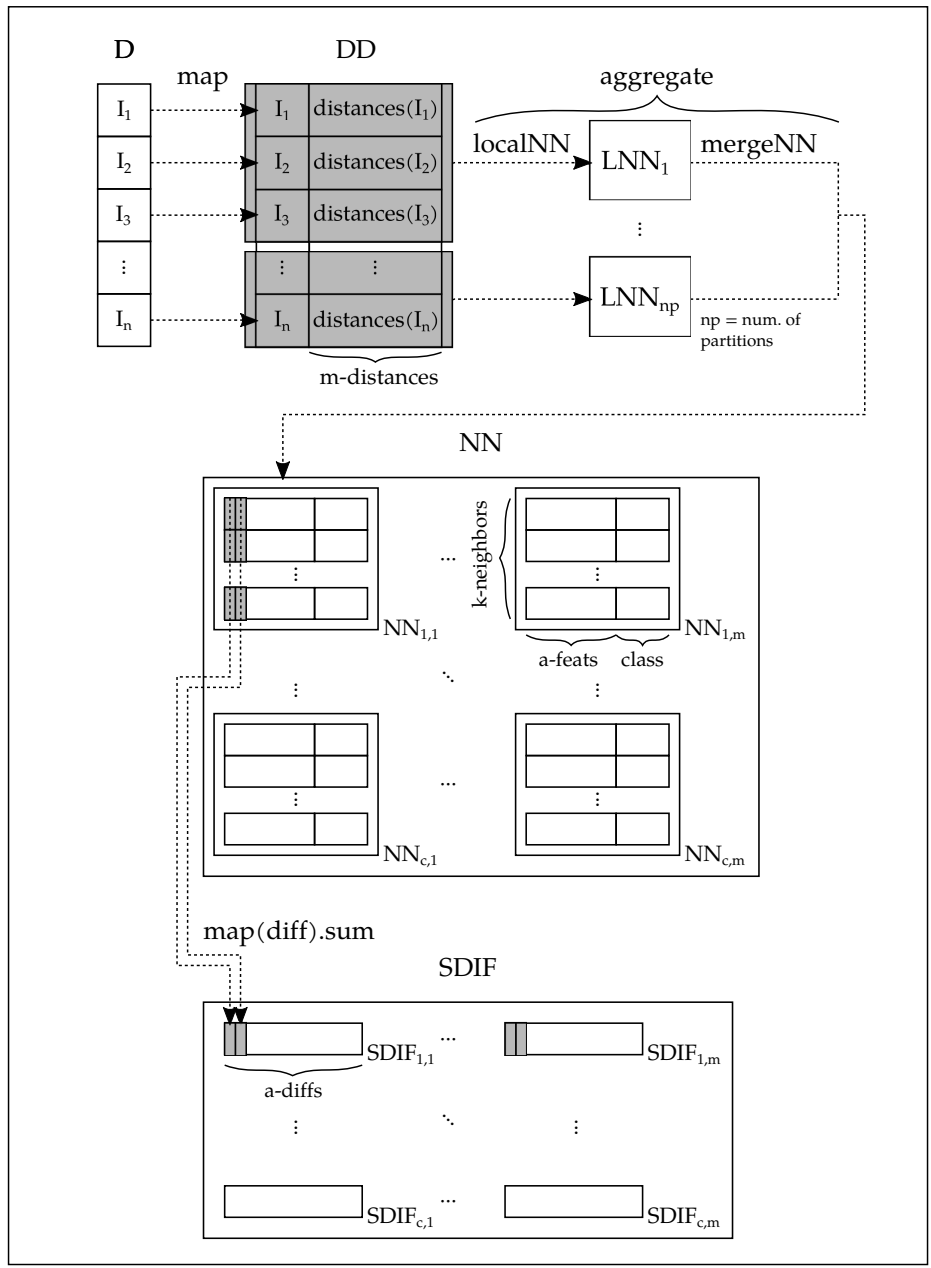}
\caption{DiReliefF's Main Pipeline}
\label{fig:relieffMain}
\end{figure}

Once the $NN$ matrix is stored on the driver program, operations are not distributed in the cluster anymore. However, this is not a problem but a requirement, because $NN$ matrix is small: $c \times m$. The last remarkable step, consists in obtaining the matrix $SDIF$. Each element $SDIF_{C,i}$ represents an $a$-size vector, and each element of this vector stores the sum of the differences for the $A$-esim feature between the $NN_{C,i}$ group of neighbors and the $R_i$ sample. Moreover, each element of the vector $SDIF_{C,i}$ can be calculated by mapping the $\mathit{diff}$ function over the $A$-esim feature of all the instances in the $NN_{C,i}$ vector and then summing this differences. Observe that this $map$ and $sum$ functions are not RDD-related anymore, but local equivalents that can be parallelized only on the driver's local threads. Finally, note that each element of each vector of the matrix $SDIF$ effectively represents the value shown in~(\ref{eq:sumOfDiffs}), thus, the final vector of weights, $W$, can be easily calculated by applying the formula given on~(\ref{eq:weightA}) using the already obtained $P$ set with the prior probabilities. Equations \ref{eq:sumOfDiffsMatrix} depict the above discussion and a resume of the algorithm's main pipeline discussed in this section is shown in Figure \ref{fig:relieffMain}.

\begin{align}
\label{eq:sumOfDiffsMatrix}
SDIF_{C,i} & = (\mathit{sdiffs}(1,R_i,C),\cdots,\mathit{sdiffs}(a,R_i,C)) \nonumber \\
f(N) & = diff(A,N,R_i) \nonumber \\
SDIF_{C,i,A} & = \mathit{sdiffs}(A,R_i,C) = NN_{C,i}.map(f).sum / k \nonumber \\
SDIF & = \begin{bmatrix}
SDIF_{1,1} & \cdots  & SDIF_{1,m}\\ 
\vdots & \ddots  & \vdots \\ 
SDIF_{c,1} & \cdots & SDIF_{c,m} 
\end{bmatrix} \nonumber \\
\end{align}

Finally, there is one implementation issue worth mentioning. As previously stated, an RDD is designed to be stored in memory, but this does not happen automatically, so, if the RDD is going to be used in future operations it must be explicitly cached. In our case, the most complex part of the algorithm is the calculation of the distances $DD$ and the $NN$ matrices. This calculation is effectively performed in a single pass through the dataset initiated by the $aggregate$ action, and therefore, caching of any intermediate result would indeed cause a waste of resources. However, the initial part of the algorithm that requires the calculations of the maximum, minimum and priors, each require a pass through the dataset $DS$ and therefore can take some advantage of caching it but only when it fits in the distributed memory of the cluster. When it does not, caching does not help because its benefits are overshadowed due to the time needed for writing the dataset to disk. As a result, in our implementation caching is disabled by default but can be enabled with a parameter (it should be enabled only when we can assure that the dataset fits in the distributed memory).

\section{Experimental Evaluation}
\label{sec:experiments}

In this section, experimental results obtained from different executions of the proposed algorithm are presented. The experiments were performed with the aim of testing the algorithm scalability and its time and memory consumption with respect with the traditional version. Tests were also performed in order to observe sample sizes were the algorithm's weights become stable.

For the realization of the tests, an 8-node cluster of virtual machines was used, one node is used as a master and the rest are slaves, the cluster runs over the following hardware-software configuration:

\begin{itemize}
\item Host Processors: 16 X Dual-Core AMD Opteron 8216
\item Host Memory: 128 GB DDR2
\item Host Storage: 7 TB SATA 2 Hard Disk
\item Host Operating System: CentoOS 7
\item Hypervisor: KVM (qemu-kvm 1.5.3)
\item Guests Processors: 2
\item Guests Memory: 16 GB
\item Guests Storage:  500 GB
\item Java version: OpenJDK 1.8
\item Spark version: 1.6.1
\item HDFS version: 2.6.4
\item WEKA version: 3.8
\end{itemize}

During the first part of the tests, the ECBDL14~\cite{Bacardit2012} dataset was used, this dataset comes from the Protein Structure Prediction field, and it was used during the ECBLD14 Big Data Competition of the GECCO'2014 international conference. The dataset has approximately 33.6 million instances, 631 attributes, 2 classes, 98\% of negative examples and occupies about 56GB of disk space. In a second part of the experiments, we use other three datasets briefly described in Table~\ref{tb:datasets}. HIGGS~\cite{Baldi2014}, from the UCI Machine Learning Repository~\cite{Lichman2013}, is a recent dataset that represents a classification problem that distinguishes between a signal process which produces Higgs bosons and a background process which does not. KDDCUP99~\cite{Ma2009} represents data from network connections and classifies them between normal connections and different types of attacks (a multi class problem). Finally, EPSILON is an artificial data set built for the Pascal Large Scale Learning Challenge in 2008\footnote{\url{http://largescale.ml.tu-berlin.de/about/}}.

\begin{table}[tbp]
\caption{Datasets used in the experiments}
\label{tb:datasets}
\begin{tabular}{lp{0.7in}p{0.5in}p{1.0in}l}
\hline\noalign{\smallskip}
Dataset  & No. of Inst.      & No. of Feats. & Features Types            & Problem Type \\
\noalign{\smallskip}\hline\noalign{\smallskip}
ECBDL14  & $\sim$33.6 million & 632            & Numerical and Categorical & Binary       \\
HIGGS    & 11 million         & 28             & Numerical                 & Binary       \\
KDDCUP99 & $\sim$5 million    & 42             & Numerical and Categorical & Multiclass   \\
EPSILON  & 1/2 million        & 2,000          & Numerical                 & Binary       \\ 
\noalign{\smallskip}\hline
\end{tabular}
\end{table}

Initially, all tests are run with a number of neighbors $k=10$ which is a typical choice \cite{Kononenko1994}, and a relatively low number of samples $m=10$ to keep execution times reasonable. However, during the stability tests larger number of samples are used. In addition, HDFS is used to store all the datasets or samples of datasets in the experiments related to the distributed version. Conversely, the local file system is used for tests of the traditional version of ReliefF.

Regarding the rest of the implementation parameters, two of them were left fixed for all the experiments. The first one selects the original $\mathit{diff}$ function in~(\ref{eq:diffNumOrig}) for the distance evaluation, as this is the one used in the ReliefF implementation in WEKA. The second parameter refers to whether or not apply caching in Spark and as stated before, caching provides benefits only when the entire datasets fully fits in memory. In order to prevent making the results obscure, it was decided that caching was disabled for all the experiments.

An important configuration issue refers to driver memory consumption. In Spark computation model, there is no communication between tasks, so all the task's results will be sent to the driver, this is especially important for the $aggregate$ action, because every task performing the $localNN$ operation will return a $LNN$ matrix to the driver that then is going to be merged. The Spark configuration parameter $spark.driver.maxResultSize$ has a default value of 1GB but it was set to 6 for all the experiments performed. This is specifically important for tests involving larger matrix sizes, i.e., those with higher values of $m$ or $c$.

\subsection{Empirical Complexity}
\label{subsec:complexity}

\begin{figure*}
  \includegraphics[width=1\textwidth]{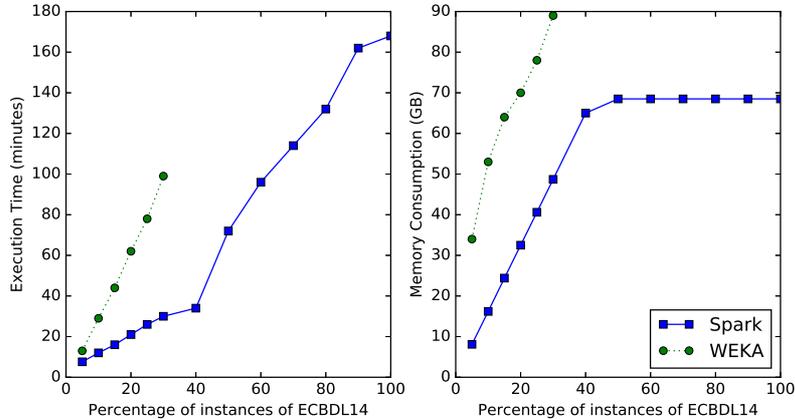}
\caption{Execution time and memory consumption of Spark DiRelieF and WEKA ReliefF versions}
\label{fig:timeAndMemory}
\end{figure*}

Figure \ref{fig:timeAndMemory} shows time and memory consumption behavior of the distributed version of ReliefF version versus the one implemented in the WEKA platform~\cite{Hall2009} for incrementally sized samples of the ECBDL14 dataset. To make the comparison possible, the WEKA version was executed under the host environment with no virtualization. It is worth noting that for the WEKA version a 30\% sample was the largest that could be tested because larger samples would need more memory than is available in the system (showing the lack of scalability). The distributed version, in addition to being able to handle the whole dataset, preserves a linear behavior in relation to the number of instances, and it is also capable of processing data in less time by leveraging the cluster nodes. Another fact to observe is the change in the slope of the linear behavior observed by the Spark version between the 40\% and 50\% samples of the dataset. This is due to the fact that during this interval the dataset overflows the available memory in the cluster and the Spark engine starts using disk storage. In other words, the algorithm is capable of maintaining a linear complexity even when the dataset does not fit into memory.

The mentioned overflow can be observed on the right graph in Figure~\ref{fig:timeAndMemory}. This graph shows a previously mentioned advantage of Spark, i.e, it is designed to run in commodity hardware. A simple look to the memory consumption of the traditional version shows that the required amount of memory quickly overgrows beyond the limits of an ordinary computer. However, in our distributed version, we can observe that using nodes, in our case with 16GB of memory, is enough to handle the task.

\begin{figure*}
  \includegraphics[width=1\textwidth]{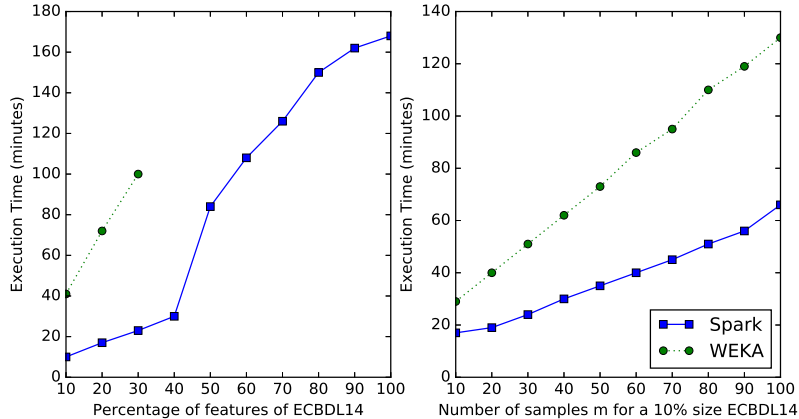}
\caption{Execution time of Spark DiReliefF and WEKA ReliefF with respect to parameters $a$ and $m$}
\label{fig:featsAndSampleSize}
\end{figure*}

Analogous run time results are obtained by varying the number of features and the number of samples (see Figure~\ref{fig:featsAndSampleSize}) confirming an empirical complexity equivalent to the original one, i.e., $\mathcal{O}(m \cdot n \cdot a)$.

\subsection{Scalability}
\label{subsec:scalability}

Now, we delve into the topic of scalability. To keep execution times within manageable limits, the largest test was performed with a 30\% sample size of the ECBDL14 dataset, following this, 10\% and 1\% sample sizes were used. Also, in oder to examine a smaller dataset, tests with 50\% and 10\% samples of the HIGGS dataset were run.

\begin{figure*}
  \includegraphics[width=1\textwidth]{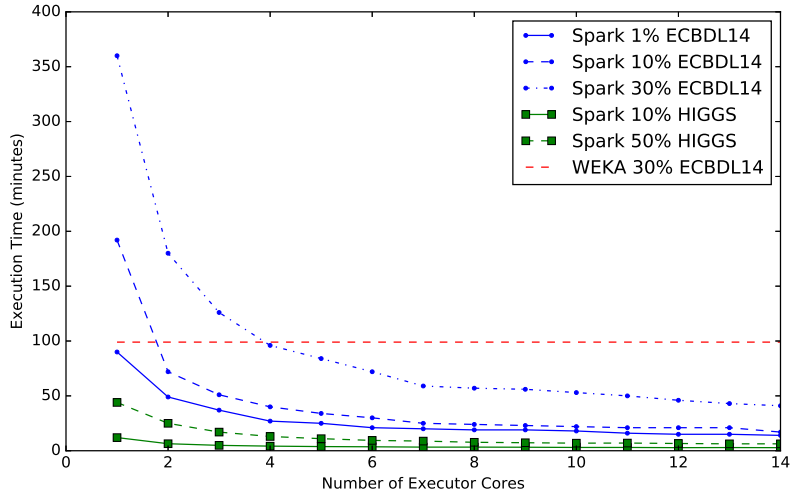}
\caption{Execution time of Spark DiReliefF and WEKA ReliefF with respect to the number of cores involved}
\label{fig:scalability}
\end{figure*}

Figure~\ref{fig:scalability} shows the behavior of the distributed algorithm with respect to the number of cores used. As it can be observed, adding cores at the beginning greatly contributes to reducing the execution time, but once a threshold is reached, the contribution is rather subtle or even null. Such threshold depends on the size of the dataset, so, larger datasets can take  benefit more of larger number of cores. On the other hand, smaller datasets will face the case were they do not have enough partitions to be distributed over all the cluster nodes. In this latter case, adding more nodes will not provide any performance improvement. As an example to quantify this fact, it can be observed in Figure~\ref{fig:scalability} that the 30\% sized sample of the ECBDL14 dataset can take advantage of 7 cores, while the 1\% sample size only can take advantage of 4 cores. Similarly, there is no practical advantage of using more than 1 core for the 10\% sample size of the HIGGS dataset.

Figure~\ref{fig:scalability} also shows with an horizontal line, the WEKA version time of the 30\% sample of the ECBDL14 dataset, since it can only take advantage of one core, the execution time is constant. However, the time is better than the distributed version in the case where 4 or less cores are involved, clearly because it does not need to deal with the driver scheduling, the selection of an executor and communication between both of them over the network, as the distributed version does. For this reason, the distributed version of ReliefF is only useful for large datasets, and in fact, regarding execution time, the 30\% sample of ECBDL14 is the only dataset in our tests that can take real advantage of the distributed version.

\subsection{Stability}
\label{subsec:stability}

The following set of tests was made in order to check the stability of the algorithm. In this case, with stability we refer to similarity of the algorithm rankings for different executions in the same dataset.

Many stability measures have been defined, a commonly used is the Consistency Index presented by Kuncheva~\cite{Kuncheva2007}, but it cannot be directly applied because it requires to define of a threshold for the selection of a subset. In addition, Kalousis et al.~\cite{Kalousis2006} proposed the use of the Pearson correlation coefficient to measure the similarity between features rankings, but previous tests showed that paradoxically returns unstable results for the EPSILON and HIGGS (pure numerical) datasets. For these reasons, we opted here for a simpler stability indicator, a mere average of the absolute differences between every feature weights of two rankings. More formally, if we have two feature ranking vectors $W1$ and $W2$, the average difference between them is calculated as shown in~(\ref{eq:stability}). 

\begin{align}
\label{eq:stability}
W1 & = (w1_1,w1_2,\cdots,w1_a) \nonumber \\
W2 & = (w2_1,w2_2,\cdots,w2_a) \nonumber \\
\mathit{AvgDiff} & = \frac{1}{a}\sum_{A=1}^{a} \left | w1_A - w2_A \right |
\end{align}

\begin{figure*}
  \includegraphics[width=1\textwidth]{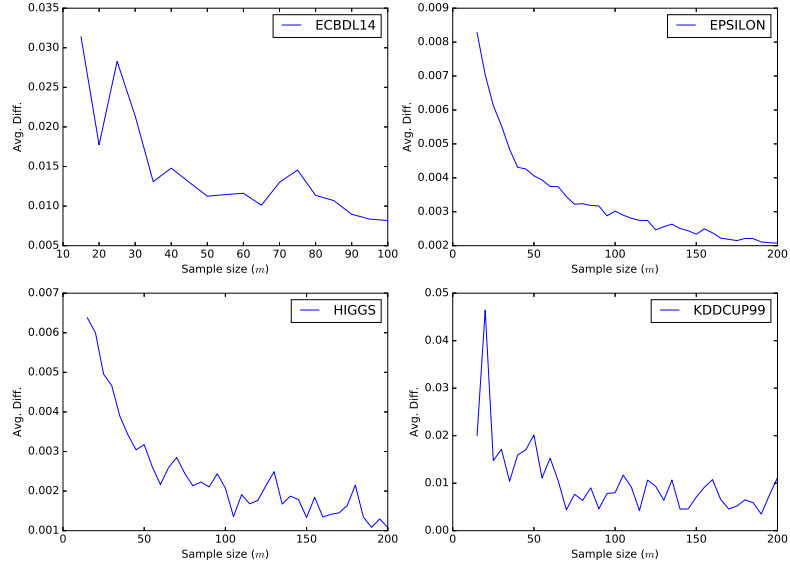}
\caption{DiReliefF's average difference in weight ranks for increasing values of $m$ in different datasets}
\label{fig:stability}
\end{figure*}

As Robnik and Kononenko~\cite{Robnik2003} stated, the correct sample size $m$ is problem dependent, so for the tests, we used all of the datasets described in Table~\ref{tb:datasets}. Moreover, their experiments show that as the number of examples increases, the required sample size diminishes. Figure~\ref{fig:stability} shows the results obtained and evidences that the biggest gain in stability is obtained with a sample size between 50 and 100 instances, thereby confirming that relative small sample sizes are enough to obtain stable results in the tested large scale datasets.

\section{Conclusions}

In this work, we have presented DiRelieF, a distributed version of the well-known ReliefF feature selection  algorithm. This version was implemented using the emerging Apache Spark programming model to deal with current Big Data requirements such as such as failure recovery and scalability over a cluster of commodity computers. Even when the ReliefF algorithm is easily parallelizable by associating jobs to each independent iteration and then merging this results, this method ties the number of jobs to the sample size (number of iterations) and requires and equal number of passes through the dataset. For this reason we designed an alternative version whose core is based on the Spark $map$ and $aggregate$ operations and on the use of binary heaps. This method does not suffer from these problems and requires a single pass through the whole dataset to perform the main operations of the algorithm compared to the $m$ passes required by the original version.

As part of the experimental work, we have also compared our proposal with a non distributed version of the algorithm implemented on the WEKA platform. Our results showed that the non distributed version is poorly scalable, i.e., it is unable to handle large datasets due to memory requirements. Conversely, our version is fully scalable and provides better execution times and memory usage when dealing with very large datasets. Our experiments also showed that the algorithm is capable of returning stable results with sample sizes that are much smaller than the size of the complete dataset.

\bibliographystyle{spmpsci}      

\end{document}